\documentclass[10pt,conference]{IEEEtran}

\usepackage{cite}

\ifCLASSINFOpdf
   \usepackage[pdftex]{graphicx}
   \graphicspath{{figs/}}
   \DeclareGraphicsExtensions{.pdf,.jpeg,.png}
\else
   \usepackage[dvips]{graphicx}
   \graphicspath{{../figs/}}
   \DeclareGraphicsExtensions{.eps}
\fi
\usepackage[cmex10]{amsmath}
\usepackage{amsthm}

\usepackage{algorithmic}

\usepackage{array}

\ifCLASSOPTIONcompsoc
  \usepackage[caption=false,font=normalsize,labelfont=sf,textfont=sf]{subfig}
\else
  \usepackage[caption=false,font=footnotesize]{subfig}
\fi
\usepackage{url}

\usepackage{enumitem}
\usepackage{xspace}
\usepackage{color, soul}
\usepackage{amsmath}
\usepackage{dblfloatfix}
\newcommand{\systemname}{\emph{ARPOV}\xspace}

\newcommand{\revision}[1]{{\color{black}{#1}}}

\newif\iffinal
\finaltrue
\newcommand{\cmtid}{139}

\iffinal
\else
\usepackage[switch]{lineno}
\fi

\begin{document}
%
\title{\systemname: Expanding Visualization of Object Detection in AR with Panoramic Mosaic Stitching}


\iffinal

\author{\IEEEauthorblockN{Erin McGowan}
\IEEEauthorblockA{New York University\\
erin.mcgowan@nyu.edu\\
}
\and
\IEEEauthorblockN{Ethan Brewer}
\IEEEauthorblockA{Spectral Sciences, Inc.\\
ebrewer@spectral.com}
\and
\IEEEauthorblockN{Claudio Silva}
\IEEEauthorblockA{New York University\\
csilva@nyu.edu}}


%

\else
  \author{SIBGRAPI Paper ID: \cmtid \\ }
  \linenumbers
\fi

\maketitle
\let\thefootnote\relax\footnote{\\979-8-3503-7603-6/24/\$31.00~\copyright~2024 IEEE\hfill}


\begin{abstract}
As the uses of augmented reality (AR) become more complex and widely available, AR applications will increasingly incorporate intelligent features that require developers to understand the user's behavior and surrounding environment (e.g. an intelligent assistant). Such applications rely on video captured by an AR headset, which often contains disjointed camera movement with a limited field of view that cannot capture the full scope of what the user sees at any given time. Moreover, standard methods of visualizing object detection model outputs are limited to capturing objects within a single frame and timestep, and therefore fail to capture the temporal and spatial context that is often necessary for various domain applications. We propose \systemname, an interactive visual analytics tool for analyzing object detection model outputs tailored to video captured by an AR headset that maximizes user understanding of model performance. The proposed tool leverages panorama stitching to expand the view of the environment while automatically filtering undesirable frames, and includes interactive features that facilitate object detection model debugging. \systemname was designed as part of a collaboration between visualization researchers and machine learning and AR experts; we validate our design choices through interviews with 5 domain experts.
\end{abstract}


\IEEEpeerreviewmaketitle

\section{Introduction}
\label{sec:intro} 

Applications of augmented reality (AR) are wide-ranging, from medicine to manufacturing. These applications have increased in both complexity and prevalence due to hardware innovations and the commercial availability of AR headsets. We have reached a point in the development of AR technology where building intelligence into AR systems is not only possible, but increasingly expected. 


Training and implementing an object detection model (ODM) on video data is a crucial step for an expanding variety of intelligent AR applications (e.g. an AI-assisted task guidance system). However, the standard method for visualizing these model outputs - bounding boxes overtop of video - lacks the spatiotemporal context necessary for thorough identification and explanation of unexpected model behavior. This approach can only display one timestep at a time, and boxes can flash quickly across the screen if detections are not consistent. Visualization of spatial context is also limited by the camera’s field of view (FoV). If the video data is collected from a head-mounted device such as an AR headset, the camera FoV will likely be much narrower than the performer’s FoV at any given point in time. It is also difficult for a viewer to situate these video frames within the spatial context of the entire scene, or know where each object was last captured by the camera. 
\par
Panorama stitching can help solve this problem. In this paper, we focus on the benefit of applying panorama stitching techniques to video frames in the context of visualizing ODM outputs captured by an AR headset. 

\textbf{Our Approach.} We propose Augmented Reality Panoramic Output Visualizer, or \systemname, an interactive tool for analyzing object detection outputs in spatial and temporal context with features tailored to the unique challenges posed by egocentric video. The \systemname interface comprises six main components: the Annotated Range Slider denotes points of interest in ODM outputs. The Timeline View summarizes model confidence, intersection over union values, detection classifications, and distances traveled by each object over a selected time. The Video Player displays the raw video. The Panorama Construction Menu enables users to control panorama building parameters and generate a panorama of the scene using selected frames, which is then displayed in the Panorama View. Finally, the Visualization Menu enables users to select various styles and filters for object detection visualization which are overlaid on the Panorama View. These views afford AR/ML experts greater spatial and temporal context for model behavior as well as knowledge of entity history (provenance), enabling improved post-hoc analysis of computer vision (CV) performance. 

The primary \textbf{contributions} of this paper are:
\begin{enumerate}
   \item \systemname, an interactive visual analytics tool with features for identifying and explaining unexpected behavior of an ODM, as well as interactive features for visualizing and interpreting the state of the environment at large, the movement of the camera, and the movement of objects within that environment.
   \item A novel technique for improving panoramic mosaic quality by clustering the frames' corresponding homography matrices by their singular values and filtering outliers. 
\end{enumerate}


\section{Related Work} 
\label{sec:relwork}

\subsection{The Need for Object Detection in AR}
Fast-paced developments in CV and AR have enabled integrated systems to continuously sense and analyze environments in real time. Intelligent mapping of a 3D space in AR requires identifying and tracking the location of objects of interest within the user's FoV. Deep learning ODMs have provided this capability across several AR contexts in the past several years \cite{ghasemi_objdet_2022}. This detection task can be divided into one or more of a few categories such as face detection, pose detection, and the identification of various inanimate objects, each of which have been used in a diverse range of applications including manufacturing \cite{kastner_humanassist_2021}, medical microscopy \cite{waithe_celldet_2020}, and assistive technologies \cite{advanti_grocery_2017,fuchs_fooddet_2020}.

Outside of related tasks such as object classification, image captioning, and segmentation (pixel level classification), the bounding box is the most common method to visualize object detection in CV \cite{xiao_detreview_2020}. Tracking of the same object over time is visualized through box permanence frame to frame, relying on the object to stay within the camera's FoV \cite{lei_tracking_2022}. Lines may be added to show object location history or future path, but only to the extent the FoV allows \cite{wang_tracking_2021}.



\subsection{Image Stitching}
Panoramic image stitching is an active research area. One approach involves feature matching and transformation estimation. Szeliski and Shum \cite{szeliski1997creating} proposed a method for automatic alignment and stitching of overlapping images using feature detection and matching. Notably, Brown and Lowe \cite{brown2007automatic} introduced an image stitching algorithm using Lowe's SIFT feature detector \cite{lowe2004sift}. Many applications, including CV library OpenCV\cite{opencv_library}, use this for image stitching. Other works have used deep learning to address certain challenges of multi-view image stitching \cite{10.1007/s11063-023-11226-z}. 


Castelo et al. \cite{castelo_argus_2023} introduced a method for expanding the FoV of an AR headset using panoramas. We build upon this concept by creating a standalone interactive tool for exploring object detection outputs with panoramic mosaics. However, our task poses several unique challenges that diverge from earlier applications. We focus on egocentric video data wherein a user may interact with the environment, creating a scene crowded with moving objects and haphazard camera panning motions. Earlier work on panoramic mosaics has either excluded moving objects in the scene \cite{HSU200481} or only dealt with simple, slow camera panning \cite{steedly2005efficiently}. 







\section{Design Requirements}
\label{sec:background}

These requirements were collected during multiple interviews with a small group of machine learning and visualization experts who all currently develop AR applications with ODMs. 

\begin{enumerate}[start=1,label={[\bfseries R\arabic*]}]

\item \label{req:model_performance} \textbf{Detailed visualizations of ODM performance}: Experts stressed the need to both summarize ODM behavior over entire sessions and investigate points of interest. 

\item \label{req:spatial_context} \textbf{Spatial context for model performance:} ODM failures in AR environments are often highly correlated with events in physical space, so spatial context is critical to model debugging. This is challenging to capture due to both camera motion inherent to egocentric video and the limited FoV typical of AR headsets.  

\item \label{req:temporal_context} \textbf{Temporal context for model performance.} Experts need to identify and understand consistent patterns in model behavior over time, ideally in a single view. While true for model debugging in general, this is especially crucial for AR applications, which rely on consistent, accurate model performance to function.   

\item \label{req:flag_POI} \textbf{Automatic flagging of potential points of interest:} AR systems can be incredibly complex, and ODM failures can have a variety of causes. Automatic detection of events that may cause interesting or unexpected model behavior (e.g. new objects, duplicate objects) enables more efficient debugging workflows. 

\end{enumerate}

\begin{figure}[!b]
\includegraphics[width=\columnwidth]{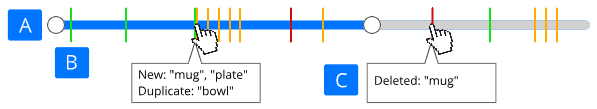}
\caption{The Annotated Range Slider, including (A) range selection for the start and end time of stitching window; (B) tick marks denoting frames in which a new object label has been detected (green), frames in which the same object label has been detected multiple times (orange), and times when a previously detected object label has not been detected for a specified count of frames (red); (C) on-hover text boxes with tick mark details.}
\label{fig:annotatedSlider}
\end{figure}

\begin{figure*}[btp]
\includegraphics[width=\linewidth]{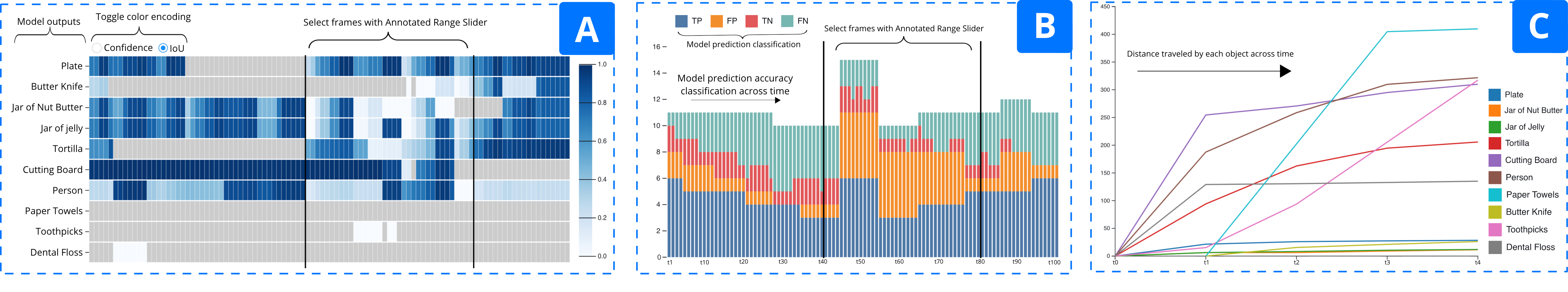}
\caption{The Timeline View allows users to toggle between three displays: (A) the Summary Matrix View, which displays model prediction confidence and the intersection over union (IoU) of predicted bounding boxes with ground truth bounding boxes; (B) the Detection Classification View, \revision{with the counts} of true positive, true negative, false positive, and false negative predictions for each frame; (C) the Distance View, which shows the distance between the centroids of consecutive predicted bounding boxes for each object adjusted within a panorama of selected frames.}
\label{fig:timelineView}
\end{figure*}

\section{ARPOV}
\label{sec:system}

\systemname is a visual analytics tool that enables troubleshooting of object detection results, with features tailored to analyzing object detection (ground truth and predicted bounding boxes) performed on RGB videos, such as those captured by current AR headsets. This video must contain multiple views of the same scene but need not be captured by a stereoscopic camera. The following sections describe the primary components of the \systemname interface, all of which are linked and interactive. All code is available on GitHub \cite{arpov_2024}.

\subsection{Annotated Range Slider}
\label{subsec:slider}

The Annotated Range Slider allows users to scrub through the video and select a range of frames. Each ODM output is matched to the nearest (in time) frame, and the slider is annotated with interactive tick marks highlighting potential points of interest (see Fig. \ref{fig:annotatedSlider}) \ref{req:flag_POI}.

\subsection{Timeline View}
\label{subsec:timeline-view}

The Timeline View (see Fig. \ref{fig:timelineView}) shows ODM performance and detected objects' status over time. It comprises three visualizations: the Summary Matrix View includes a heatmap wherein each pixel corresponds to a frame/object pair. Pixel color intensity denotes either model confidence for that object at that frame, or the intersection over union (IoU) of the predicted and ground truth bounding boxes; the user can toggle between these  \ref{req:model_performance} \ref{req:temporal_context}. 

The Detection Classification View includes a stacked bar chart with true positive, true negative, false positive, and false negative prediction counts for each frame. Together, these views allow users to explore model performance and identify areas where  confidence and accuracy are low \ref{req:model_performance} \ref{req:temporal_context}. 

The user can select frames of interest in the Summary Matrix and Detection Classification Views using the Annotated Range Slider (see Section~\ref{subsec:slider}), denoted in both by vertical bars, and generate a panorama of selected frames (see Section~\ref{subsec:panoramaView}). This updates the Distance View, which displays the distance between the centroids of consecutive predicted bounding boxes for each object adjusted within the panorama. This shows when objects are in motion, and how far they travel each time they move \ref{req:model_performance} \ref{req:spatial_context} \ref{req:temporal_context}. 


\begin{figure*}[h]
\includegraphics[width=\linewidth]{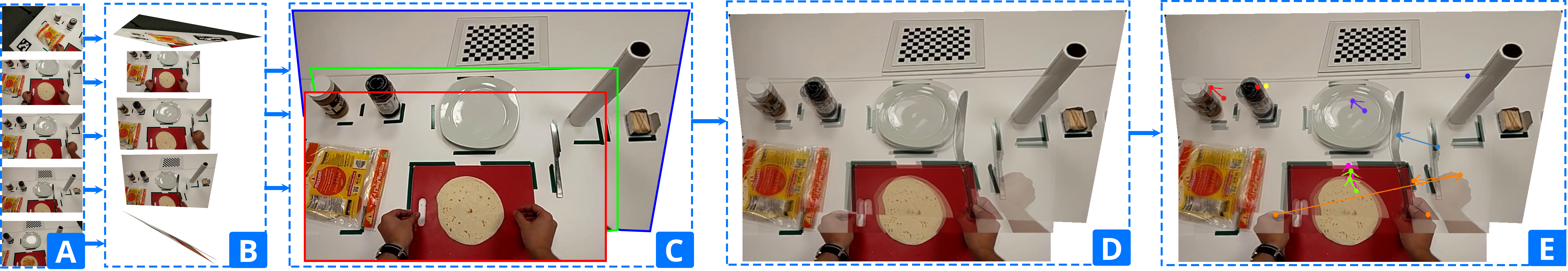}
\caption{The panorama construction pipeline takes in raw frames (A), computes and filters their homographies (B), positions (C) and composites (D) them on the canvas and overlays corresponding ODM outputs (E).}
\label{fig:panoramaConstructionPipeline}
\end{figure*}

\begin{figure*}[!t]
\includegraphics[width=\linewidth]{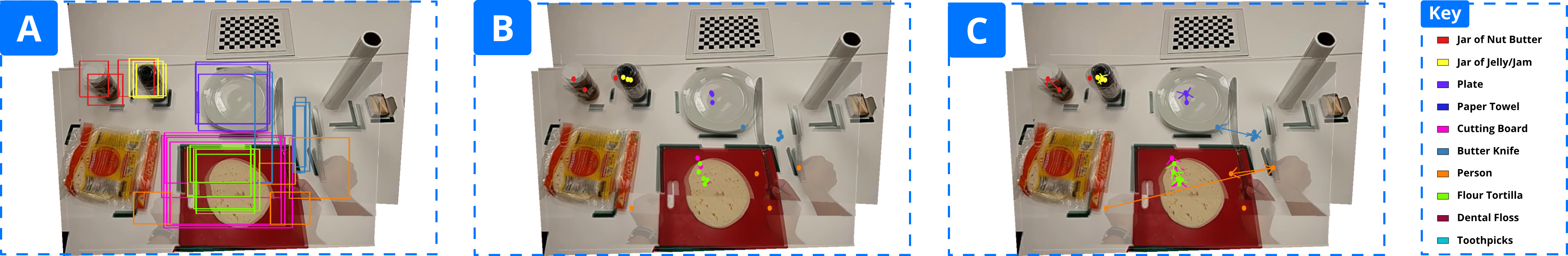}
\caption{\systemname enables users to view ODM outputs in three different visualization styles: (A) Bounding Boxes, (B) Centroids, and (C) Arrows. Each color denotes a different object label, as shown in the key.}
\label{fig:annotationStyles}
\end{figure*}

\subsection{Panorama View}
\label{subsec:panoramaView}

The Panorama View is a unique component of \systemname that allows users to simultaneously explore the spatial and temporal context of selected video frames and model output data by generating panoramic mosaic visualizations \ref{req:model_performance} \ref{req:spatial_context} \ref{req:temporal_context}. 

\textbf{Panorama Construction.} The Panorama Construction Menu allows users to specify panorama stitching parameters. We project selected frames onto the plane of a base frame (defaults to median) using a \textit{homography matrix}, which applies a series of geometric transformations. To compute the homography for each frame, we first compute its keypoints and descriptors, which denote distinctive features. Many feature detectors compute keypoints and descriptors, each with their own advantages and disadvantages \cite{8346440}. To accommodate these trade-offs, \systemname allows users to select from popular feature detectors chosen for speed and accuracy: BRISK \cite{brisk}, ORB \cite{orb}, KAZE \cite{kaze}, and AKAZE \cite{akaze}.

Next, we match feature pairs between the base frame and each of the other selected frames. However, not every pair will be correctly matched to the same point in space. This is especially true for AR headset data; wearer and object movement can cause motion blur, obscuring features. Changing lighting conditions can also lower matching accuracy. Thus we employ a threefold process to filter out incorrect matches. First, we apply Lowe’s ratio test \cite{lowe2004sift}, filtering matches without sufficient distance between the best and second-best matches. Next, we compute a homography matrix between each frame and the base frame by direct linear transformation \cite{agarwal2005survey} such that the back-projection error is minimized. For frame pairs with over 4 feature matches, we take many random subsets of 4 non-collinear corresponding point pairs, estimate the homography using each subset, and compute homography quality using the number of RANSAC \cite{Fischler_Bolles_1981} inliers. Each homography is refined with the Levenberg-Marquardt method \cite{Levenberg_1944, Marquardt_1963} to further reduce reprojection error. Users can adjust the Lowe's ratio test and RANSAC reprojection thresholds.

To position each frame in the panorama, we take the normalized dot product of its homography and the heterogeneous coordinates of its corners. We then use Three.js \cite{threejs} to load frames as textures and render them on a WebGL canvas. The user can adjust the alpha value (opacity) of these textures both before and after panorama generation. See Figure \ref{fig:panoramaConstructionPipeline} for an example of this panorama construction process. 

\textbf{Visualizing Object Detection.} ODMs typically output bounding boxes in a two-plus-coordinate format. To place these in the panorama, we must apply the same geometric transformations (homography) applied to their corresponding frame. We then compute each box's centroid. For each distinct object label, we calculate pixel distance between each consecutive adjusted centroid detection. These distances are reflected in the Distance View of the Timeline View (see Section \ref{subsec:timeline-view}). 

We visualize object detection outputs by rendering them onto a transparent texture that lies overtop the panorama on the WebGL canvas. The user can select one of three visualization styles (see Figure~\ref{fig:annotationStyles}). The first, \textquotedblleft Bounding Boxes," draws standard rectangular bounding boxes based on the two coordinates provided in the ODM output. The second, \textquotedblleft Centroids," draws only the centroid of each bounding box. This eliminates clutter and facilitates identification of clusters  indicating that an object stayed in one place for a relatively long period of time. The third, \textquotedblleft Arrows," draws arrows connecting consecutive centroids for a given object label over time, facilitating movement tracking. If an object label is detected multiple times in the same frame, each instance is assigned to the closest group of arrows for that label. 

Mark color denotes object label. Chosen colors are bright, unnatural, and stand out against most environments. The user can filter outputs by  confidence and object label, and also highlight the outline of a specific frame in the panorama via slider. Note the panorama need not be regenerated each time the user selects new object visualization options. 
 
\subsection{Homography Filtering}
\label{subsec:homographFiltering}

The existing methods we use to filter incorrect feature matches are not always sufficient, especially in a busy environment full of objects and unpredictable, shifting lighting conditions. \systemname introduces a process which eliminates frames with incorrect matches by aspects of their homography matrices that indicate either 1) the frame has been stretched abnormally within the panorama or 2) the frame has been flipped (highly unlikely with a head-mounted camera). 

\textbf{Removing Abnormally Stretched Frames.} We define the intrinsic camera parameters matrix $K$, which relate a given camera to an ideal pinhole-camera model \cite{Apple_Developer_Documentation}.
To account for deviation from this ideal, we compute $H^{\prime} = KHK^{-1}$ for each homography $H$ \cite{Malis_Vargas_2007}. We will use this calibrated homography to compare scaling of each frame in the $x$ and $y$ directions. To isolate a measure of this scale transformation, we decompose each $H^{\prime}$ using singular value decomposition, yielding $H^{\prime} = U \Sigma V^T$. We are concerned with $\Sigma$, a 3x3 matrix whose diagonal values represent the singular values of $H^{\prime}$. The first two singular values, $\sigma_1$ and $\sigma_2$, correspond to relative scaling in the $x$ and $y$ directions. 

\begin{figure*}[!t]
\includegraphics[width=\linewidth]{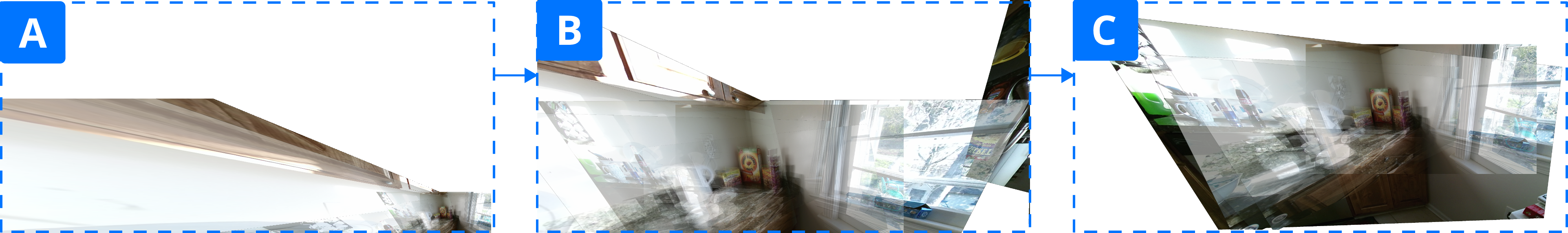}
\caption{The \systemname homography filtering feature removes frames causing distortion in the panoramic mosaic. Shown are (A) a panorama generated using out-of-the-box OpenCV functions, (B) the same panorama with frames stretched far out of proportion with the rest of the mosaic removed, and (C) the same panorama with vertically and horizontally flipped frames removed.}
\label{fig:case_study_1}
\end{figure*}

We identify outliers via k-means clustering the frames by their $\sigma_1$ and $\sigma_2$ values. We select the number of clusters, $k$, by calculating the Within-Cluster-Sum of Squared Errors $WSS$ for different values of $k$ and choosing the value of $k$ at which $WSS$ sharply drops (mimicking the graphical Elbow Method \cite{cui2020introduction}). Finally, we compute the distance between the centroid of each cluster and the singular values of the base frame (the origin, since the base frame's homography is the identity matrix). We remove any frames not contained in the cluster closest to the origin (see Figure~\ref{fig:case_study_1}\revision{)}.

\textbf{Removing Flipped Frames.} We apply each frame's homography to the heterogeneous coordinates of its corners, computing $(x_1, y_1)$, $(x_2, y_2)$, $(x_3, y_3)$, and $(x_4, y_4)$ (clockwise from top left). We check the following: (1) $y_3 < y_2$; (2) $y_4 < y_1$; (3) $x_2 < x_1$; (4) $x_3 < x_4$. If both (1) and (2) are true, the frame has been vertically flipped. If both (3) and (4) are true, the frame has been horizontally flipped. If either (1) or (2) or either (3) or (4) are true, the frame has become twisted so that it converges to a single point at some location. We remove these from the panorama (see Figure~\ref{fig:case_study_1}). 

\begin{figure*}[!t]
\includegraphics[width=\linewidth]{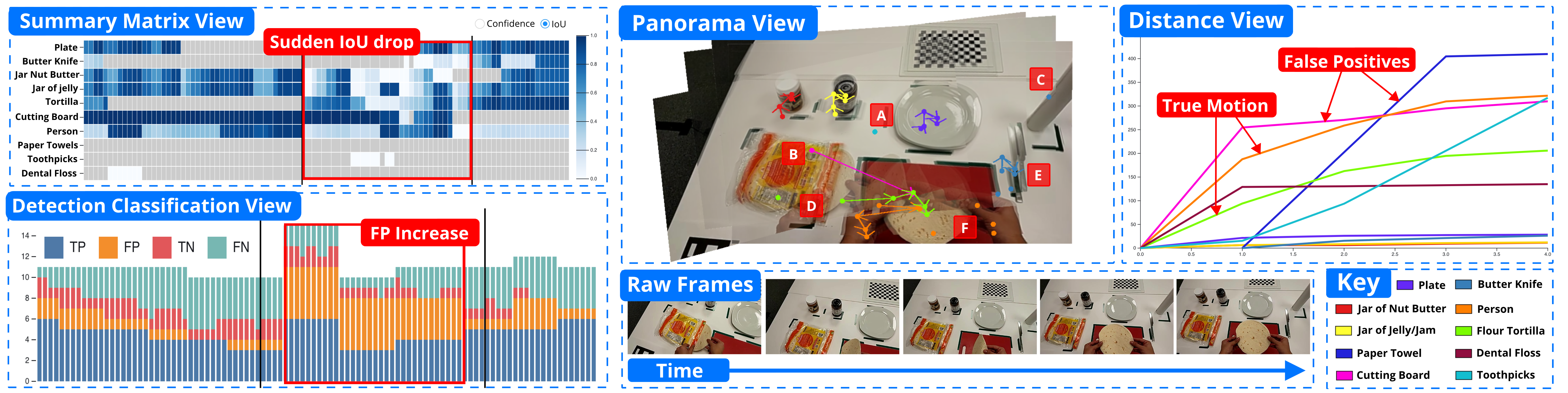}
\caption{A panoramic mosaic of 5 frames sampled from a selection based on Summary Matrix and Detection Classification views. Shown are (A) a falsely detected object that is not actually present in the scene (``toothpicks"); (B) An object detected at the wrong location (``cutting board"); (C) An object the model failed to detect (``paper towels"); (D) Correctly detected duplicate objects (``flour tortilla"); (E) An object that was incorrectly detected as a duplicate (``butter knife"); (F) An object that was correctly detected with duplicates at some timesteps, but with an incorrect amount of duplicates (``person").}
\label{fig:case_study_2}
\end{figure*}

\section{Case Study}
\label{sec:casestudies}

\subsection{Data}
\label{subsection:case_study_data}

We explore a use case for \systemname showcasing distinctive issues that may arise with AR headset data. Additional examples can be found on GitHub \cite{arpov_2024}. The video was captured by a HoloLens~2 mixed-reality headset \cite{hololensresearchmode} while a person performed a recipe task. We focus on the front-facing camera, which captures RGB8 video at 15 fps at a 760×428 resolution. We generated bounding boxes using Detic \cite{zhou2022detecting}, an ODM which classifies objects using a given vocabulary (see Fig.~\ref{fig:case_study_2}), in real time at $\sim$ 2 fps. We generated "ground truth" bounding boxes for each frame with Language Segment-Anything (LangSAM\revision{)} \cite{githubGitHubLucamedeiroslangsegmentanything}
\revision{Since we apply} LangSAM to every frame, it follows that these bounding boxes will be more accurate than the closest DETIC output for most frames, due to the lower DETIC frame rate used to enable real-time detection.

\subsection{Analyzing Object Detection Model Outputs}
\label{subsection:casestudy2}

We present a scenario in which an AR guidance system developer uses \systemname to identify reasons for model failure. The developer first surveys the Summary Matrix View for a recipe video (see Section \ref{subsection:case_study_data}), observing that the IoU values for multiple objects suddenly drop (see Figure~\ref{fig:case_study_2}). The Detection Class View also shows a significant increase in false positive detections around that time. She selects these frames using the Annotated Range Slider and generates a panorama. 

The panorama shows two objects in motion during this portion of the recording: the tortilla and the performer's hands (see Figure \ref{fig:case_study_2}). By selecting the Arrow visualization style (see Figure \ref{fig:annotationStyles}), she confirms that the model captured this motion. She makes several observations regarding model detection successes, failures, and duplicate handling (see Figure~\ref{fig:case_study_2}A-F).

The developer notes that only two objects seem to have moved in the Panorama View, but the Distance View shows the cutting board, paper towels, toothpicks, and dental floss in motion during the selected frames as well. This is due to the false positive detections. Between the Distance View and Panorama View, the developer can easily discern which objects are truly in motion and how far they traveled. 

\section{Evaluation}
\label{sec:evaluation}

\vspace{-0.15cm}
\subsection{Expert Interviews}
\label{subsec:expert-interviews}

We interviewed 5 domain experts: 2 Research Engineers (E1, E2) and 3 Postdoctoral Researchers (E3, E4, E5), all with 5-10 years experience with ODMs and 3-10 years experience in AR applications. Each interview lasted 25-40 minutes. We presented \systemname, demonstrated the scenario described in Section~\ref{subsection:casestudy2}, and answered questions. Then we discussed the experts' initial impressions of the tool, its functionalities and features, and its potential application to their own workflows. 

\subsection{Expert Feedback}
\label{subsec:expert-feedback}

Expert feedback was highly positive, demonstrating interest in using \systemname for their tasks and suggesting enhancements. Notably:

\textbf{E1} was interested in using \systemname for model debugging, noting \textquotedblleft I like the fact that you put the trajectory of the objects....because it can help us to understand, perhaps, why the model failed....It is a good indication." 

\textbf{E2} concluded \textquotedblleft I think the most valuable part of the tool [is] the ability to browse through the video and apply homography on demand....This brings it into this singular interactive visualization space, especially where you can take a screenshot and share it. It's quite valuable." E2 also suggested system layout improvements that were made prior to publication.

\textbf{E3} called the Timeline View \textquotedblleft a highlight" of the system, and expressed interest in integrating it into his own workflow for an AR application that utilizes ODMs. E3 also emphasized the benefit of the views being linked. 

\textbf{E4} commented \textquotedblleft The panorama seems really cool. The fact that you can even get a sense of things like head movement, and how everything connects together something that I, personally had never even been able to appreciate," and \textquotedblleft [the Distance View] is very transparent with the ability to understand what [the object detector] may be doing right or doing wrong."

\textbf{E5} noted that confusion matrices provide a summary \textquotedblleft but [the Timeline View] gives you the details, almost like a higher-resolution peek into what’s going on....For me, this would immediately point me to which part of the video is causing the model to fail....You want that granularity when you're trying to debug something,” and \textquotedblleft I think [the Distance View] is a very novel feature here....This idea, of visualizing the distance, I think that's very ingenious. It's useful." 

Multiple experts believed the tool seems well-suited to optical flow visualizations; we note this in future work. 


\section{Conclusion}
\label{sec:conclusion}

We presented \systemname, an interactive visualization tool that enables AR application developers to efficiently visualize and analyze ODM behavior (a critical component of many AR applications) using panoramic mosaics. 

\textbf{Limitations.} Visual clutter and computation time will scale with the number of frames included in each panorama. The chosen visualization style may also exacerbate this issue. 

\textbf{Future Work.} We intend to add tracking and optical flow visualizations. We also plan to explore how the tool could be expanded to accommodate real-time analysis and other (AR and non-AR) applications which rely on object detection.

\section{Acknowledgements}
This work was supported by the DARPA PTG program. Any opinions, findings, and conclusions or recommendations expressed in this material are those of the authors and do not necessarily reflect the views of DARPA.
Erin McGowan was funded by an NYU Tandon Future Leader Fellowship.

\bibliographystyle{IEEEtran}
\bibliography{example}
%
%



\end{document}